# BICLUSTERING READINGS AND MANUSCRIPTS VIA NON-NEGATIVE MATRIX FACTORIZATION, WITH APPLICATION TO THE TEXT OF JUDE

JOEY MCCOLLUM AND STEPHEN BROWN

ABSTRACT. The text-critical practice of grouping witnesses into families or texttypes often faces two obstacles: Contamination in the manuscript tradition, and co-dependence in identifying characteristic readings and manuscripts. We introduce non-negative matrix factorization (NMF) as a simple, unsupervised, and efficient way to cluster large numbers of manuscripts and readings simultaneously while summarizing contamination using an easy-to-interpret mixture model. We apply this method to an extensive collation of the New Testament epistle of Jude and show that the resulting clusters correspond to human-identified textual families from existing research.

## 1. INTRODUCTION

Genealogical analysis has had a prominent role in New Testament (NT) text-critical theory even before it was popularized in the work of Westcott and Hort [1]. Indeed, one of the steps in their approach, that of classifying manuscripts (MSS) into families and texttypes based on their shared readings, goes back over a century-and-a-half earlier to the works of Mill, Bentley, and Bengel [2]. In theory, the rationale for this is that the more two MSS agree in their readings, the more likely they are to represent a close common ancestor or to have an exemplar-copy relationship themselves; the goal is that by grouping witnesses in this way, the critic can then weigh them according to how purely they represent their group or how derivative they are from common sources.

The theory is not without obstacles, however. Despite the emphasis they placed on the genealogical method, Westcott and Hort misused the method by overlooking the effects of *contamination*, or mixture of readings characteristic of different text-types, on the genealogy they were attempting to derive [3]. It turns out that such mixture is somewhat ubiquitous. Indeed, as more MSS are discovered and studied, placing them into families and texttypes by hard assignment only blurs the lines between these groups even more.

An additional complication in the assignment of MSS to groups is the dual problem of assigning readings to groups. Any two witnesses will probably agree in a majority of their readings, so simply counting places of agreement is insufficient [4, 5]. In order to classify texts into well-defined, well-separated groups, we must determine which readings are the most significant for this purpose. So we must first determine the readings that the most characteristic MSS of a group share and that few or no other MSS share. But in order for us to do this, the MSS must be assigned to groups already. This leaves us with a problem of *co-dependence*: Characteristic MSS of a given type are determined by which characteristic readings they have, and characteristic readings of a given type are determined by which characteristic MSS attest to them.

These observations have spurred the increased use of approaches that exchange the assumptions of texttype theory for other assumptions. Seeing the benefits of these methods, some researchers have started to question the continued relevance of methods based on texttypes. Others have proposed to abandon the underlying theory altogether [6].

Yet the idea of texttypes has not been rejected universally. Epp, for one, has argued for its continued value [2]. More generally, the assumptions made by other methods introduce limitations of their own, some of which the theory of texttypes can overcome. It is also important to recognize that texttype-based methods and newer approaches are not mutually exclusive, but can be used in conjunction to achieve more refined results. We will elaborate on these points briefly. The main implication of our work with regard to theory, however, is that contamination and co-dependence are not truly insurmountable obstacles to texttype-based methods; as we will show, both issues can, in fact, be handled.

In this paper, we present non-negative matrix factorization (NMF) as a simple, unsupervised, and efficient computer-based method for isolating clusters of MSS and their most important readings simultaneously. It is a pre-genealogical method only, in the sense that it does not infer any directed relationships among readings, texts, or groups. As such, it is not intended to replace more complex genealogical methods, but to help researchers generate hypotheses that these methods can test and refine. In the first section, we review known classification methods with their advantages and disadvantages. In the second section, we show how NMF addresses the problems of contamination and co-dependence and discuss its advantages over other methods. (This section involves some mathematics, but the more technical details can be found in the references.) In the third section, we describe our application of NMF to Tommy Wasserman's full collation of 560 MSS of the epistle of Jude [7]. In the last section, we show that the use of NMF in different settings yields results in the form of recognizable texttypes and families established in the literature.

## 2. State of the Art

The history of textual criticism has seen the development of numerous classification methods. We will not attempt an exhaustive treatment here, but instead focus our attention on the following generalizations:

(1) *Base-text*: This method, used in NT textual criticism at its earliest stages, classifies MS texts based on their deviations from a single base of readings. The base could be anything, but historically, it was the Textus Receptus (TR). Classes were formed based on their common agreements against the base text.

(2) *Quantitative*: This method, introduced by Colwell [8], compares the text of each MS to that of every other MS. Similarity is generally measured as a simple count or proportion of readings at which two texts agree. The set of pairwise similarity measures is then used to group texts according to a variety of algorithms [9, 10].

(3) *Profile*: The Claremont Profile Method (CPM), developed by McReynolds and Wisse [11, 12], improves the efficiency and classification power of existing methods by taking identified groups and determining which patterns of readings provide the best "profile" for each group. These profiles can then be used to determine the most likely group of an unclassified text and the strength of a classified text's group membership.

(4) *Factor Analysis*: This method, which has been developed and put to use extensively at Andrews University [13, 14, 15], first determines the *factors*, or combinations of readings that are most correlated to one another among MS texts, and then groups MSS by their strongest factors. In this way, it combines the "text-side" approach of (2) with the "reading-side" approach of (3).

(5) *Stemmatic*: Stemmatic methods aim for a more precise and structured classification of texts by examining changes in every variation unit between extant texts and nodes representing their potential ancestors. The goal is to construct an undirected stemma of *maximum parsimony*, or the fewest changed readings along all links of the stemma. Examples of these methods can be found in the recent work of Spencer, Wachtel, and Howe [16, 17].

(6) *Coherence-based*: The Coherence-Based Genealogical Method (CBGM), developed by Mink [18], improves (5) by producing directed stemmata that are robust to contamination. It starts by constructing a directed *local stemma* for the readings at each variation unit based on general agreement among readings' witnesses and transcriptional probability, so that readings have prior and posterior relationships. Then, it constructs a directed local stemma for the texts at each unit based on their general agreement and their relative proportions of prior and posterior readings where they disagree. These first two steps are applied iteratively to account for the co-dependent relationship of "good" readings and "good" texts. Finally, the local stemmata of texts are merged to form a *global stemma* that optimizes the objective of simplicity (i.e., as few ancestors to a text as possible) under the constraint that the directed relationships of all readings are preserved. A useful introduction can be found in Wachtel's recent feature on the method [19].

We will summarize each method's advantages and disadvantages below.

Method (1) is simple to understand, but it has many shortcomings. It requires the assumption of a base text, and it only defines groups in terms of their disagreements with this text, when their agreements might also be informative. Because it offers no procedure for deciding which readings are the most signficant in determining groups, it also fails to address the problem of co-dependence between readings and texts. Typically, the evaluation of which readings are important is done under human supervision and can be slow for a large number of readings. This method can account for contamination, but only after groups have been determined. A brief discussion of this method can be found in Ehrman's essay on classification methods [5].

Method (2) presents easy-to-interpret results, requires no human supervision, and dispenses with the unnecessary assumption of a base text. Its major disadvantage is its computational cost: Building a table of pairwise similarity measures for $n$ MSS requires $(n^2 - n)/2$ comparisons, which becomes inconvenient for a large number of MSS. Because it counts readings rather than weighing them, it fails to address the problem of co-dependence. Typical clustering algorithms do not accommodate contamination.

Method (3) is much more efficient by design than (2), produces results that are simple to understand, and can identify and quantify contamination between groups in a MS's text. Its greatest weakness is that it does not identify the groups to be profiled, but assumes that they are known.[1] It therefore does not even attempt to address the problem of co-dependence. Finally, despite its efficiency, it is a supervised method, so it will not be as fast as some computer-based methods.

---

[1] CPM has, in fact, been criticized on the basis of its application with poorly-identified groups [20]. Because of this, it is best used in conjunction with methods like (2) or (4) [5].

Method (4) is extremely efficient, is unsupervised, and does not require the assumption of a base text. As such, it could reasonably be said to supersede method (2) as a quantitative method. Furthermore, the way the method fits readings and texts together in factors may be considered an adequate approach to the problem of co-dependence. The drawback is that the results of factor analysis can be difficult to interpret: The method often assigns a negative coefficient for one or more groups to a text, and there is no obvious sense of what that means. The problem is that factor analysis does not provide a model that accounts for additive mixture between groups. Naturally, this becomes an obstacle to identifying contamination. For these reasons, factor analysis is often refined using method (3) [13, p. 37].

Method (5) is unsupervised, and it makes no assumptions about a base text. Furthermore, unlike methods (1)–(4), it avoids controversy over the use of texttypes by focusing instead on transmissional details at the level of individual texts and readings. The price of this level of detail, of course, is complexity in implementation: The construction of an optimal stemma is a computationally expensive task, so one must sacrifice accuracy to complete it in an acceptable amount of time. The other major disadvantage is that simple stemmata cannot account for contamination. In addition, simple stemmatic approaches make no statements about "good" texts or readings until one has to orient the stemma.

Method (6) accounts for the problems of contamination and co-dependence between texts and readings without requiring any assumptions about underlying texttypes. In other areas, however, it is lacking. First, it is a supervised method. Second, between the initialization of a pairwise MS coherence table and the supervised construction of local stemmata, the method is very time-consuming. Third, and perhaps most importantly, it requires the assumption of a working initial text, which for all practical purposes is tantamount to the base-text assumption of method (1). This assumption severely limits the scope of what can be achieved: CBGM can help refine existing hypotheses, but it cannot compare significantly different hypotheses. CBGM has risen to prominence relatively recently, having only been applied on a large scale to the catholic epistles and Acts. During that time, it has seen discussion as well as criticism.[2]

As we will show, NMF combines many of best qualities of methods (1)–(6). It does not rely on a base text. It is unsupervised and extremely fast, even with data as large as Wasserman's Jude collation. It performs biclustering on MSS and readings at the same time, so it does not need a profile method to be applied on top of it. Most importantly, NMF finds an equilibrium in the co-dependent relationship of "good" witnesses and "good" readings, and it does so in a way that is robust to contamination between groups. As NMF models mixture additively, its results are easy to interpret. See Table 1 for a comparison of methods (1)–(6) and NMF.

Of course, NMF relies on the assumptions of texttype theory, so it is subject to the criticisms of that theory. Even so, while a comprehensive discussion of texttype theory and its relevance is beyond the scope of this paper, it should be noted that the results of NMF can still be used to benefit other methods in the pre-processing stage. CBGM and other stemmatic methods, for instance, are often applied only to a sample of all available MSS and readings for the sake of time and computational resources; NMF can provide a smart selection of significant MSS and readings in this situation.

---

[2]For more information, see Houghton's survey [21] , Wasserman's overview [22], and *TC*'s featured articles on CBGM at rosetta.reltech.org/TC/v20/index.html.

Table 1. Comparison of textual classification methods. The "A-text?" column label refers to the assumption of a base or working initial text, the "Texttypes?" label to the assumption of underlying texttypes, the "Unsup.?" label to the question of whether the method is unsupervised, the "Fast?" label to whether the method is efficient, and the "Contam.?" and "Co-dep.?" labels to whether the method is robust to contamination and co-dependence, respectively.

|  | A-text? | Texttypes? | Unsup.? | Fast? | Contam.? | Co-dep.? |
|---|---|---|---|---|---|---|
| Base-text | ✓ | ✓ |  |  |  |  |
| Quantitative |  | ✓ | ✓ |  |  |  |
| Profile |  | ✓ |  | ✓ | ✓ |  |
| Factor Analysis |  | ✓ | ✓ | ✓ |  | ✓ |
| Stemmatic |  |  | ✓ |  |  |  |
| Coherence-based | ✓ |  |  |  | ✓ | ✓ |
| NMF |  | ✓ | ✓ | ✓ | ✓ | ✓ |

## 3. Theoretical Basis

NMF was popularized by Daniel D. Lee and H. Sebastian Seung in their 1999 paper [23]. Since then, it has found use in a wide array of fields (see [24] for a detailed survey). With regard to fields most relevant to textual criticism, it has found use in text mining for the purposes of document clustering and topic modeling [25] and in computational biology for the purposes of classifying gene expressions in DNA microarrays [26] and more recently for determining biological admixture, or ancestry, coefficients [27]. The following treatment of the mathematical basis of NMF is primarily a summary, but the interested reader is encouraged to refer to [28] for details.

Suppose that we have a collation consisting of $m$ readings (with readings in the same variation unit treated as distinct objects) and $n$ MSS. A natural way to represent this collation is an $m \times n$ matrix, where the rows represent the readings and the columns represent MSS. A cell at row $i$ and column $j$ contains a 1 if MS $j$ attests to reading $i$, and a 0 otherwise. We will call this collation matrix $\mathbf{X}$. A small example of such a matrix is shown in Fig. 1.

$$
\begin{array}{r}
\texttt{ms.1} \quad \texttt{ms.2} \quad \texttt{ms.3} \quad \texttt{ms.4} \\
\begin{array}{r}
\texttt{unit.1.reading.1} \\
\texttt{unit.1.reading.2} \\
\texttt{unit.2.reading.1} \\
\texttt{unit.2.reading.2} \\
\texttt{unit.2.reading.3} \\
\texttt{unit.3.reading.1} \\
\texttt{unit.3.reading.2}
\end{array}
\left[
\begin{array}{cccc}
1 & 1 & 0 & 0 \\
0 & 0 & 1 & 1 \\
1 & 0 & 1 & 0 \\
0 & 1 & 0 & 0 \\
0 & 0 & 0 & 1 \\
1 & 1 & 1 & 0 \\
0 & 0 & 0 & 1
\end{array}
\right]
\end{array}
$$

Figure 1. Matrix representation of part of a collation. As expected, each column (MS) only has a nonzero value in a single reading in a given unit.

We want to find $k$ latent features underlying this data. In our case, the features are texttypes or textual families; the larger $k$ is, the finer the groupings are. For the purposes of this application, we assume that the observed readings and MSS have been generated by these hidden features. Fig. 2 gives an illustration of this model.

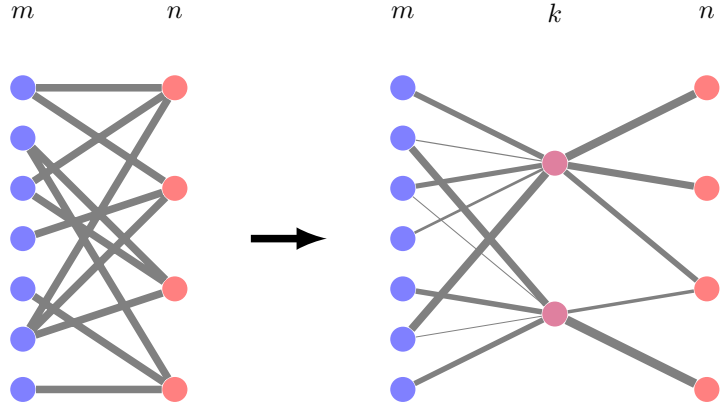

FIGURE 2. A graphical illustration of the latent feature model assumed by NMF, using the data from Fig. 1 and $k = 2$. Notice how the post-factorization model weighs readings by the information they provide about a cluster and accounts for a mixture of these clusters in `ms.3`.

The corresponding equation for this generative process is

$$\mathbf{X} \approx \mathbf{WH}. \tag{1}$$

Here, $\mathbf{W}$, called the *basis* matrix, is an $m \times k$ matrix describing the makeup of each group in terms of linear combinations of readings, and $\mathbf{H}$, called the *coefficient* or *mixture* matrix, is a $k \times n$ matrix describing the makeup of each MSS's text in terms of linear combinations of groups. So our goal is to find group membership coefficients for readings and for texts that, when combined, explain the data as best as possible. Since the data matrix $\mathbf{X}$ is obviously non-negative and we want to describe it in terms of "sums of parts," we restrict $\mathbf{W}$ and $\mathbf{H}$ to be non-negative, as well. This model has several advantages over standard clustering models: First, it allows readings and texts to belong to more than one cluster, which is critical for providing group profiles and dealing with contamination; second, it assigns weights to readings so that the ones more characteristic of a single cluster have higher priority than those shared among several; and third, it assigns weights to MSS, so we can see which ones are the strongest and purest representatives of their groups.

The "closeness" of the product $\mathbf{WH}$ to $\mathbf{X}$ can be measured in a variety of ways, but in this paper, we will use

$$||\mathbf{X} - \mathbf{WH}||_F^2, \tag{2}$$

which is the squared *Frobenius norm*, or the total sum of squared differences between each cell of $\mathbf{X}$ and the corresponding cell of $\mathbf{WH}$. So our goal in any factorization will be to minimize this quantity.

How, then, do we find $\mathbf{W}$ and $\mathbf{H}$? We must first choose initial matrices $\mathbf{W}_1$ and $\mathbf{H}_1$ ourselves, either with random positive values in each cell or with an educated guess. Then we alternate between updating them using the following rules:

$$(3) \qquad \mathbf{W}_{t+1} = \arg\min_{\mathbf{W} \geq 0} ||\mathbf{X} - \mathbf{W}\mathbf{H}_t||_F^2$$

$$(4) \qquad \mathbf{H}_{t+1} = \arg\min_{\mathbf{H} \geq 0} ||\mathbf{X} - \mathbf{W}_{t+1}\mathbf{H}||_F^2$$

In other words, for each matrix in the factorization, we fix the values of the other matrix and choose new non-negative values for this matrix that minimize the value of function (2). Clearly, this means that the value of function (2) will never increase from one step to the next, but can only decrease or remain the same. Until we reach a point in the process where rules (3) and (4) no longer change anything, we iteratively update $\mathbf{W}$ using the current $\mathbf{H}$ and then update $\mathbf{H}$ using the new $\mathbf{W}$. Our aim is that by repeating these alternating optimizations, we will eventually reach a fixed point in the process where function (2) is no longer improved by these updates.

At this point, the reader may recognize the lurking shadow of co-dependence between readings ($\mathbf{W}$) and witnesses ($\mathbf{H}$). How do we know that the loop of updates won't reach a stable point before the objective function does? And what happens if the process never reaches a fixed point? As it turns out, such scenarios can never happen:

**Theorem** ([29, 28]). *Any limit point of the sequence $\{\mathbf{W}_t, \mathbf{H}_t\}_{t=1}^{\infty}$ generated by rules (3) and (4) is a stationary point of function (2).*

From this theorem, we are guaranteed that our update process will not reach a stable point without the objective function doing the same. If we add any upper bound (it can be as large as we need to ensure accuracy) to the entries of $\mathbf{W}$ and $\mathbf{H}$, then we can also guarantee that the update process has at least one limit point. So NMF with objective function (2), update rules (3) and (4), and (arbitrarily large) upper bounds on the matrix entries will always converge to a factorization that is at a stationary point of the objective function.[3]

As this analysis has shown, NMF provides a natural model for identifying textual families and more than adequately addresses the problems of contamination and co-dependence. With regard to contamination, it not only detects the degrees and sources of contamination in individual texts, but also quantifies the importance of specific readings to textual clusters. With regard to co-dependence, we have shown that NMF's use of iterative refinement of group readings and group witnesses is guaranteed not only to stop, but to stop at a critical point of the function it is trying to minimize.

## 4. Application

4.1. **Data.** We applied NMF to Tommy Wasserman's comprehensive collation of the epistle of Jude [7]. We considered this a good testing ground for the method for several reasons:

- The size of the collation, which might be prohibitive for more complex, supervised methods, can be handled efficiently and automatically by NMF.

---

[3]It should be noted, however, that the process and the objective function may have more than one stationary point, meaning that one NMF run may reach a *locally* optimal, but not *globally* optimal, factorization. It is therefore important to run NMF with good initial guesses for $\mathbf{W}$ and $\mathbf{H}$ or to repeat it many times with different random guesses.

- The collation is complete over nearly all readings and MSS.[4] We can therefore avoid any biases associated with previous selections of "genealogically significant" readings and texts. Moreover, starting with virtually all available evidence, we can discover new readings and texts of significance and add confidence to existing ones whose significance we re-discover.
- To the best of our knowledge, no other application of this scale has been done with Wasserman's work. We hope that our work will spark continued research involving his collation and inspire work towards collations of equal scale elsewhere in the NT.

The collation covers 560 MSS, including 3 papyri and 38 lectionaries, across 360 variation units. In encoding the data, we ignored lacunae, partially lacunose or uncertain readings;[5] readings from non-continuous-text sources such as correctors' hands;[6] and units contained within larger, overlapping omissions.[7] All readings that were not skipped, including omissions, were then represented by their own row in the collation matrix, with MSS represented in columns. (See Fig. 1.) The result was a $1346 \times 560$ matrix with 178887 non-zero entries.

After running NMF on this matrix in different settings, we observed that highly lacunose MSS and singular readings were being isolated in their own clusters. This likely occurred because these witnesses and readings constituted outliers with high influence on the objective function of the factorization. To account for this, we removed all singular readings from consideration and treated all MSS with fewer than 300 readings as secondary, non-continuous texts. Filtering these out, we were left with a $789 \times 518$ matrix with 155400 non-zero entries. The excluded MSS are listed on page 20.

4.2. **Prior Weights on Readings.** We applied NMF to our collation matrix with the entries being weighed uniformly on the one hand, and using the inverse document frequency (IDF) scheme on the other hand. Uniform weighting, as its name suggests, weighs all readings equally prior to NMF; all entries of the matrix are therefore 0-1 entries. IDF weighting, developed to facilitate information retrieval in the context of terms and documents [31], is a heuristic that seeks to weigh individual terms by their specificity. While a theoretical justification of its use has been elusive [32], it has proven to be of great practical effectiveness in text mining tasks.

The definition is simple. If a term $t$ occurs $n_t$ times among $n$ documents, then we assign it a weight

$$\text{(5)} \qquad \qquad \text{IDF}(t) = \log \frac{n}{n_t}.$$

Here, log is a logarithm that can be taken to any base greater than 1. With this weighting function, the more documents term $t$ appears in, the less information it provides about any specific document. The word *the* is an example of such a term. If it appears in all $n$ documents, then its IDF weight would be $\log(n/n) = \log(1) = 0$.

---

[4]Wasserman notes that his apparatus does not record the most frequent orthographic variants, such as instances of movable *nu*, final vowel elisions in prepositions and conjunctions, itacisms, and other common vowel interchanges [7, p. 129–130]. But this is actually good for our purposes, since such readings are considered unimportant for MS classification [30, p. 27–28].

[5]These readings contain underlying dots or brackets in Wasserman's apparatus.

[6]Non-continuous texts pose the same problems for NMF that they do for other methods like CBGM. Nevertheless, once clusters have been determined for the continuous-text data, we can make inferences about the non-continuous-texts on the basis of their readings. See Appendix A for more details.

[7]Another way to handle units contained in larger omissions would be to treat them as omissions themselves. We did not attempt to encode the data in this way, but the number of variation units in question is small enough that we do not expect significantly different results.

Meanwhile, a word appearing in only a single document would have the larger weight $\log(n)$.

In our case, we would take readings as "terms." The highest-specificity readings, then, would be those that occur more rarely, with singular readings receiving the highest prior weight. Balancing this out is the fact that singular readings are less informative about non-trivial clusters (i.e., those consisting of more than a single MS), which will lead NMF to give these readings a lower final weight than it gives to readings with attestation from slightly larger groups. Sometimes, however, singular readings will exert enough influence to bias the factorization towards outlying MSS. This is what initially happened with our data, as we showed in subsection 4.1. When this happens, it is prudent to perform outlier detection and removal, as we have done. Ultimately, in maximizing the cohesiveness of MS clusters, non-singular readings that are shared exclusively by the MSS in a given cluster are likely to become the most important readings. Meanwhile, readings that are common to multiple clusters will be given less weight and therefore will play less of a role in NMF.

Each of these two weighting schemes is best suited for a different set of tasks. Uniform weighting is better in the context of clustering on the "whole picture" of readings and explaining as much of the variance between MSS as possible. This is applicable, for example, to the construction of a text-critical apparatus, where we want to group many MSS under a few sigla for the purpose of succinctness, while having to list as few exceptional cluster members as possible.[8] IDF weighting is more useful in producing weighted profiles of textual families based on their most exclusive readings. In general, it is more aligned with human intuition in identifying textual groups, as we will see.

4.3. **Implementation.** For ease of use, we stored all apparatus matrices, along with their row and column headers, as Microsoft Excel spreadsheets. For all computational work, we used release 3.5 of the Python programming language.[9] To read and write data from and to Excel spreadsheets, we used the Python `pandas` package.[10] For running NMF on our data, we chose to use `nimfa`, an open-source Python library [33]. This library best fit our needs because it offered a variety of NMF versions, including versions having the convergence guarantees summarized in section 3, numerous factor initialization methods, and factorization quality measures. For most of our data manipulation needs, including linear algebra for matrix operations, we used the SciPy stack of open-source Python modules for scientific computing.[11] To factor our collation data in `nimfa`, we used the `Lsnmf` method, an implementation of the alternating least-squares formulation of NMF proposed by Lin [28]. We ran this with values of $k$ ranging from 2 to 8 and a maximum iteration limit of 8000. We used a single run in each case, seeding it with NNDSVD initialization, a non-random initialization method that has been empirically shown to result in faster, lower-error factorizations [34]. We found that single runs initialized in this way matched the objective function values for hundreds to thousands of runs with random initialization, and in fact tended to give sparser factors.

This implementation of NMF was run separately on a platform with an Intel i7-4770 quad-core processor and 16GB of memory, which we will denote `P1`, and a platform with an Intel 2 Duo dual-core processor and 2GB of memory, which we will denote `P2`. The differences in performance between the two platforms will be detailed in the following sections.

---

[8]Of course, the more contaminated the scribal tradition is, the more exceptions one can expect to see in the apparatus. But even then, the level of compression achieved for the average reading may still outweigh the number of exceptions.

[9]`python.org/`

[10]`pandas.pydata.org/`

[11]`scipy.org/`

## 5. Results

5.1. **Uniform-weighted results.** Table 2 gives summary statistics for the uniform-weight NMF runs and results for $2 \leq k \leq 8$. In general, NMF obtained factorizations that explained much of the variance in the observed data, and it did so in a very short time.

TABLE 2. Summary statistics for uniform-weight NMF results. Here, `n_iter` gives the number of iterations before convergence, `P1.time` and `P2.time` give the running time in seconds on both platforms, `dist` gives the value of the objective function from equation (2), `evar` gives the explained variance as a proportion between 0 and 1, and `W.spar` and `H.spar` give sparseness measures between 0 and 1 according to Hoyer's formula [35].

| $k$ | n_iter | P1.time | P2.time | dist | evar | **W.spar** | **H.spar** |
|---|---|---|---|---|---|---|---|
| 2 | 119 | 0.9380 | 2.7512 | 8686.7455 | 0.9496 | 0.3407 | 0.7045 |
| 3 | 613 | 3.3250 | 12.0516 | 8209.1803 | 0.9524 | 0.3439 | 0.6377 |
| 4 | 625 | 4.1603 | 16.9735 | 7851.3511 | 0.9545 | 0.3465 | 0.6401 |
| 5 | 1695 | 11.4074 | 45.9784 | 7597.1562 | 0.9559 | 0.3465 | 0.6583 |
| 6 | 1919 | 17.0553 | 53.5620 | 7366.8363 | 0.9573 | 0.3476 | 0.6773 |
| 7 | 2489 | 22.6352 | 81.0192 | 7165.2404 | 0.9584 | 0.3484 | 0.6851 |
| 8 | 2644 | 24.9981 | 77.5668 | 6964.6268 | 0.9596 | 0.3481 | 0.7124 |

We will now examine the results for $k = 8$ in detail. Table 9 lists the mixture coefficients for consistently-cited witnesses in the NA[28] apparatus for Jude [36]. Note that the coefficients have not been normalized. We have left them as-is in order to preserve their absolute magnitude, which we can interpret as a "confidence score" for classification. If we were to divide each coefficient in a given row by the row-wise sum, we could interpret the scaled coefficients as mixture proportions. Under such normalization, MS `01` would be interpreted as 78% cluster 1, 15% cluster 2, 4% cluster 8, and 3% cluster 7.

In order to determine the textual groups represented by the clusters, it is instructive to look at their most representative witnesses. Tables 10–17 list the top 15 MSS in each cluster. For the purposes of profiling secondary witnesses, we will also want to know the most important readings in each cluster. Tables 18–25 list these readings in order of their coefficients.

The group behind cluster 1 is perhaps the easiest to identify: This cluster represents the Alexandrian texttype. Perhaps not surprisingly, one of its best representatives is `03` (**B**), with papyrus $\mathfrak{P}^{72}$ being another leading member. The remaining top representatives include a handful of NA[28]'s consistently-cited witnesses. Uncials `01` (**ℵ**), `02` (**A**), and `04` (**C**) also fall under this cluster, but as Table 9 shows, they all have strong enough elements of mixture with other clusters that they do not make it to the top of the cluster's list. The cluster contains 59 MSS in total.

Cluster 2 appears to be a mixture of two textual families identified in existing literature: $\mathbf{f}^{1739}$ and $\mathbf{f}^{2138}$ [37, p. 1136–1163, 1166–1175]. The former group has also been identified in 2 Peter [38, p. 45–47], and in the catholic epistles, it shares important readings with the old Georgian versions [39]. Its namesake is a consistently-cited witness in NA[28]. One of `1739`'s scribes claimed to have copied it from an ancient codex, and scholars conjecture that its exemplar dates back at least to the fourth or fifth century. Further evidence for the family's antiquity has been found in its close similarity to the text used by Origen [40]. The connection with Origen has led some to posit that $\mathbf{f}^{1739}$ represents the controversial "Caesarean" texttype in the catholic epistles [39]. The latter group has also been identified in 2 Peter [38, p. 51–53], and in James, 1 Peter, and 1 John, its core members have been shown to have a connection to the Harklean Syriac [41, 42]. So $\mathbf{f}^{1739}$ and $\mathbf{f}^{2138}$ both attest to early forms of several of the catholic epistles, and the same situation likely holds in Jude, as well. The cluster is small, at 22 MSS, but the witnesses are generally cohesive. The fact that their readings overlap enough for their MSS to be grouped together also suggests that $\mathbf{f}^{1739}$ and $\mathbf{f}^{2138}$ are closely related to each other in most variation units.

Cluster 3, as its witnesses make clear, represents the group of lectionaries. The existence of a distinct lectionary textual group has been recognized for some time [43], but a thorough examination of this group in the catholic epistles was long delayed. The first, and perhaps most extensive, work in this area was done by Junack [44]. Junack's work confirmed the existence of a large and cohesive textual family among the Byzantine lectionaries. At least in the context of the epistle of Jude, our results based on Wasserman's complete collation should give additional weight to these findings. Our results also agree with Junack's identification of $\ell596$ as an exceptionally non-Byzantine lectionary; in fact, Table 10 lists it as a strong representative of the Alexandrian texttype. The cluster does not consist exclusively of lectionaries, as it contains 41 MSS total, but the non-lectionary MSS are lower on the list due to mixture.

Cluster 4 clearly represents the Byzantine subgroup $\mathbf{K}^{\mathbf{r}}$, also known as $\mathbf{f}^{35}$, as can be seen from the overlap between Table 15 and the list of collated MSS for 2 John–Jude in [45]. This cluster is by far the largest, with 172 MSS assigned to it, and it exhibits great cohesion among its purest representatives. To its disadvantage, however, it contains no witnesses dating earlier than the tenth century.

Cluster 5 appears to represent another Byzantine subgroup, but it is unclear if it corresponds to any previously-known subgroup. While not as massive as cluster 4, it is still large with 101 MSS. Perhaps the most noticeable quality is that it appears to be the earliest Byzantine subgroup. It contains the following five ninth-century MSS: 1424 ($\mathbf{H}_{5,j} = 1.3575$), 049 ($\mathbf{H}_{5,j} = 1.1871$), 018 ($\mathbf{K}$) ($\mathbf{H}_{5,j} = 1.1231$), 1862 ($\mathbf{H}_{5,j} = 0.9247$), and 1841 ($\mathbf{H}_{5,j} = 0.6001$).

Cluster 6 undoubtedly represents the textual family $\mathbf{f}^{453}$ [37, p. 1098–1101]. Its earliest witness is the tenth-century MS 307, but this same MS is also a consistently-cited witness in NA$^{28}$ (see Table 9). This group was independently identified in the catholic epistles through stemmatic methods by Spencer, Wachtel, and Howe, who noted that it "contains states of text that are thought to be important for the formation of the Byzantine text" [16]. The family is of moderate size, containing 35 MSS.

Cluster 7 also looks Byzantine, but like cluster 5, its precise identity is unclear. Like cluster 5, it seems to represent a text earlier than that of cluster 4; its earliest witnesses are dated to the tenth century, but two prominent representatives of the group are uncials 056 and 0142, and both of these possess Alexandrian elements (quantifiable as cluster 1 mixture proportions of 15% and 23%, respectively). The cluster is of moderate size, consisting of 59 MSS.

Cluster 8 appears to be von Soden's $\mathbf{K}^{\mathbf{c}}$ Byzantine subgroup [46, p. 1761], as can be seen from the presence of the following $\mathbf{K}^{\mathbf{c}}$ MSS in the cluster: 912 (von Soden's α366), 390 (δ366), 2085 (α465), 234 (δ365), 42 (α107), 1594 (δ375), 1405 (α555), 51 (δ364), 223 (α186), 1860 (α377), 97 (α260), and 421 (α259). The cluster as established by NMF has no witnesses from earlier than the tenth century, and of its purest representatives, the oldest is the eleventh-century MS 42, but the group is large enough, with 29 members, and tight-knit enough, with generally strong mixture coefficients, that its archetype is surely older than the tenth century.

There are a few observations to make here. NMF on a uniform-weight collation matrix reveals a number of distinct subgroups not only of the Alexandrian texttype, but also of the Byzantine texttype. In particular, the Byzantine texttype splits into the lectionary group, $\mathbf{K^r}$, $\mathbf{K^c}$, and two additional groups. It should be noted, therefore, that the Byzantine MSS do not form a monolithic group in Jude. Clusters 5 and 7 form two more large, as-yet unknown Byzantine families. While we might expect one of these clusters to correspond to the larger, more general $\mathbf{K}$ group, the MSS traditionally assigned to that group are divided between both clusters. It may be the case that the $\mathbf{K}$ MSS are divided in Jude, with two thirds of the family siding with the oldest MSS in the group and one third taking the other side.

A cross-reference from Tables 18–25 to Wasserman's apparatus reveals that NMF in the uniform-weight setting tends to assign higher basis coefficients to common, widely-divided readings than it does to rarer readings exclusive to groups. This is the result of NMF trying to minimize the number of misclassified readings when all readings are all equal in weight. To minimize unexplained variance, readings are chosen on how cleanly they divide the entire body of MSS into their assigned clusters. For these reasons, an "important" reading in this setting will likely represent multiple clusters, but a given cluster can be uniquely identified by patterns of readings. This, in essence, reflects the methodology of Wisse and McReynolds's original formulation of the Claremont Profile Method [11, 12]. While this has the unfortunate side-effect of not clustering readings as sparsely as we might like, it is useful for certain purposes. In particular, it allows us to identify widely-split variant units, which may represent early divisions in the scribal tradition, and to determine where different families side in these splits. Table 3 gives a short list of widely-divided readings and their support among uniform-weight NMF groups.

TABLE 3. Uniform-weight NMF cluster support for highly divided readings. Cluster-wide support is determined not by manuscript count, but by basis coefficient, with a reading (resp. group of readings) being considered representative if its coefficient (resp. sum of coefficients) is a least twice the value of every alternative's coefficient (resp. sum of coefficients) in that variation unit. The readings are primarily split between "...Ἰησοῦ Χριστοῦ..." and "...Χριστοῦ Ἰησοῦ..." in v1u4–8; "πάντα...Ἰησοῦς ἅπαξ," "πάντα...κύριος ἅπαξ," "ὑμᾶς ἅπαξ τοῦτο...κύριος," "ὑμᾶς τοῦτο ἅπαξ...κύριος," and "ἅπαξ τοῦτο...κύριος" in v5u12–20; "...Μωϋσέως..." and "...Μωσέως..." in v9u24–28; "δὶς ἀποθανόντα ἐκριζωθέντα" and "δὶς ἀποθανόντα καὶ ἐκριζωθέντα" in v12u42–46; "...ἀσεβεῖς" and "...ἀσεβεῖς αὐτῶν" in v15u14–18; and "ἐπιθυμίας ἑαυτῶν" and "ἐπιθυμίας αὐτῶν" in v16u14–16.

|  | *Alex* | $\mathbf{f^{453}}$ | $\mathbf{f^{1739\,+\,2138}}$ | *Lect* | $\mathbf{K^r}$ | $\mathbf{K^c}$ | $\mathbf{K_{2/3}}$ | $\mathbf{K_{1/3}}$ |
|---|---|---|---|---|---|---|---|---|
| v1u4–8 | I.X. | I.X. | I.X. | X.I. | I.X. | X.I. | X.I. | I.X. |
| v5u12–20 | Split | ατκ. | Split | υατκ. | υατκ. | υατκ. | υατκ. | υτακ. |
| v9u24–28 | Μωυσ. | Μωυσ. | Μωσ. | Μωσ. | Μωσ. | Μωυσ. | Split | Μωσ. |
| v12u42–46 | om. | και | om. | και | om. | om. | om. | om. |
| v15u14–18 | om. | om. | Split | om. | αυτ. | αυτ. | αυτ. | αυτ. |
| v16u14–16 | Split | αυτ. | Split | εαυτ. | εαυτ. | αυτ. | Split | αυτ. |

As an example, the Robinson-Pierpont Greek NT lists `v9u24-28` and `v16u14-16` as divided Byzantine readings and opts for Μωϋσέως and αὐτῶν, respectively, as the original readings [47]. While the $\mathbf{K}_{2/3}$ cluster is split in both cases and offers no strong evidence either way, Μωϋσέως has the support of *Alex*, $\mathbf{f}^{453}$, and $\mathbf{K^c}$, and αὐτῶν has the support of $\mathbf{f}^{453}$ and $\mathbf{K^c}$. The external evidence shows earlier and more diverse testimony in favor of Μωϋσέως, with the same situation to a lesser degree in favor of αὐτῶν.[12] Valuing diversity of testimony, we therefore agree with Robinson and Pierpont's textual decisions on the divided readings in Jude. The results of NMF may also reveal readings that have not yet been recognized as divided Byzantine readings; `v1u4-8` and `v5u12-20` are two candidates.[13]

Of course, in other applications, we would want to identify readings that are individually more exclusive to their groups. In these cases, we view less common readings as more valuable *a priori*. Thus, to get sparser results, we must turn to NMF in the IDF-weight setting.

5.2. **IDF-weighted results.** Table 4 gives summary statistics for the IDF-weight NMF runs and results for $2 \leq k \leq 8$. Because IDF weighting assigns greater importance to less common readings, NMF tends to find more exclusive bases for clusters in this setting. Another positive effect of IDF weighting is that it allows NMF to isolate more distinctive clusters, often due to differences that uniform-weight NMF overlooks. A disadvantage, as can be seen in Table 4, is that when the factorization sets aside especially rare readings for the sake of more common, cohesive ones, the high weight of the ignored readings reduces the explained variance of the model.[14] For all $k$, an NMF run in this setting took at most a couple seconds on both platforms.

Table 4. Summary statistics for IDF-weight NMF results. Here, `n_iter` gives the number of iterations before convergence, `P1.time` and `P2.time` give the running time in seconds on both platforms, `dist` gives the value of the objective function from equation (2), `evar` gives the explained variance as a proportion between 0 and 1, and $\mathbf{W}$`.spar` and $\mathbf{H}$`.spar` give sparseness measures between 0 and 1 according to Hoyer's formula [35].

| $k$ | n_iter | P1.time | P2.time | dist | evar | $\mathbf{W}$.spar | $\mathbf{H}$.spar |
|---|---|---|---|---|---|---|---|
| 2 | 0.2028 | 1.3663 | 20 | 9641.5789 | 0.1228 | 0.4625 | 0.7984 |
| 3 | 0.2028 | 1.7377 | 37 | 9389.1472 | 0.1458 | 0.5561 | 0.8237 |
| 4 | 0.3276 | 1.3492 | 37 | 9155.7879 | 0.1670 | 0.5972 | 0.8271 |
| 5 | 0.3900 | 1.8167 | 57 | 8961.6038 | 0.1847 | 0.6260 | 0.8187 |
| 6 | 0.3900 | 1.8582 | 66 | 8771.8117 | 0.2020 | 0.6326 | 0.8085 |
| 7 | 0.7176 | 2.6713 | 98 | 8600.4446 | 0.2176 | 0.6496 | 0.8132 |
| 8 | 1.0668 | 4.3561 | 188 | 8451.6279 | 0.2311 | 0.6659 | 0.8007 |

We will now examine the results for $k = 8$ in detail. Table 26 lists the mixture coefficients for consistently-cited in the NA[28] apparatus for Jude [36]. As in the uniform-weight case, no normalization has been applied to the coefficients. Tables 27–34 list the most representative MSS in each cluster, and Tables 35–42 list the most representative readings for each cluster.

---

[12]We would not consider the agreement of $\mathbf{K^r}$ and *Lect* to be especially diverse, as they both are closely related to the Byzantine texttype. The only non-Byzantine support comes from $\mathbf{f}^{1739 + 2138}$ for Μωϋσέως and $\mathbf{f}^{1739}$ for αὐτῶν (the cluster's support there is essentially split between $\mathbf{f}^{1739}$ and $\mathbf{f}^{2138}$). This relationship may be worth closer study in the future.

[13]At these locations, Robinson-Pierpont gives only the readings Ἰησοῦ Χριστοῦ δοῦλος and ὑμᾶς ἅπαξ τοῦτο ὅτι ὁ κύριος, respectively. If we were to account for the readings of the other $\mathbf{K}$ subgroups here, we would include the readings Χριστοῦ Ἰησοῦ δοῦλος and ὑμᾶς τοῦτο ἅπαξ ὅτι ὁ κύριος, respectively, in the margin.

[14]See subsection 4.2 for more details on the problem and how to address it.

Cluster 1 clearly represents $\mathbf{f}^{2138}$. We also note that now the cluster contains no representatives from $\mathbf{f}^{1739}$, and for this reason, it now consists of only 24 MSS. An important observation in Table 26 is that uncial 04 (**C**) has its highest mixture coefficient in this cluster, which suggests that it shares many characteristic readings, or at least some high-weight characteristic readings, with $\mathbf{f}^{2138}$. The first few can be found in Table 35 and are the following: `v24u18.2`, `v14u26-36.2`, `v23u2-22.17`, and `v25u24-30.1`.

Cluster 2 appears to represent the majority of the Byzantine texttype, as 318 MSS are members of it. The group appears to be cohesive enough not to be split up when $k = 8$, but its mixture coefficients are also the lowest of any cluster by far, which suggests that there is variance among the members of the cluster. This likely arises in splits between Byzantine subgroups at common readings; in the IDF setting, these readings will have low enough weight not to split the cluster, but their total weight will suffice to produce noticeable variance within the group. Nevertheless, the characteristic readings of the cluster have strong coefficients, which indicates that uniquely Byzantine readings are shared even by the texttype's different subgroups.

Cluster 3 represents $\mathbf{f}^{1739}$. The group is small, consisting of only 8 MSS, but this is the result of its being split apart from $\mathbf{f}^{2138}$. Uncial 04, while not a member of this cluster, still shares some significant readings with it. These readings include `v5u4.2`, `v22u2-10.4`, and `v14u26-32.2`.

Cluster 4 contains the Alexandrian witnesses. The cluster, which consists of 39 MSS, is slightly smaller than its counterpart in the uniform-weight setting. Due to more exclusive readings being assigned prominent places in the cluster basis, the order of the most representative witnesses has shifted somewhat. Notably, uncial 02 (**A**) and $\mathfrak{P}^{72}$ are considered slightly more "Alexandrian" than 03 (**B**) in the IDF setting. A glance at Table 26 will reveal that 02 is relatively pure in its mixture coefficients, while 01 (**ℵ**), 03, and 04 all have at least some mixture with Cluster 7. We will revisit this in a moment.

Cluster 5 represents $\mathbf{f}^{453}$. The group consists of 22 MSS here, which is a bit smaller than it was from uniform-weight NMF. Like the other clusters, its readings should now be more exclusive to the group. Little else has changed.

Cluster 6 is obviously the lectionary group. It now consists of 73 MSS, which means that it may have borrowed some MSS placed elsewhere in the uniform-weight clusters. Its readings are now more exclusive to the group, but little is different otherwise.

Cluster 7 is a curious group consisting of only 13 MSS. Most of its members were lumped under the Alexandrian cluster in the uniform-weight setting, so it appears to have some relationship with the Alexandrian text. The top two MSS, 915 and 88, demonstrate a high level of agreement in both the catholic and Pauline epistles. In the catholics, they and a few other members of this cluster (442, 621, 1243, 1846, and 2492) read δι' ὕδατος καὶ πνεύματος καὶ αἵματος in 1 John 5:6. In 1 Corinthians, 88 and 915 attest to an infamous variant that places 14:34–35 at the end of the chapter, with the only other Greek MS support coming from Western witnesses. Their support for that reading has led to much debate over whether or not they have a common source in a localized Western text and whether or not they support the theory that 1 Corinthians 14:34–35 is an interpolation [48, 49, 50]. Despite the rarity of some of its other readings, the cluster's characteristic readings are shared by the Alexandrian uncials 01, 03, and 04, and 044 (**Ψ**), which suggests that the cluster preserves some ancient readings. As the cluster itself does not appear to have been identified in the literature, we will designate it by $\mathbf{f}^{915}$ here.

Cluster 8 is another unusual group, made up of 21 MSS. Most of its members were mixtures of multiple Byzantine subgroups in the uniform-weight setting, so it appears to represent a small and distinct branch of the Byzantine texttype. Its top representatives, MSS **618**, **460**, **177**, **337**, and **1738**, are strong, pure representatives of the family. Some of the MSS themselves are noteworthy. Scrivener describes **618** as "valuable, but with many errors" [51, p. 294]; he finds a similar text in **460** and considers this MS "an important copy" [51, p. 291]. Apart from this, the cluster itself does not seem to have received much study. Lacking an existing name for it, we will designate it $\mathbf{f}^{618}$ in this paper. We observe one other connection between these MSS outside the catholic epistles, in the letter to the Romans. There, many of the family MSS contain the subscription πρὸς Ῥωμαίους ἐγράφη ἀπὸ Κορίνθου διὰ Φοίβης τῆς διακόνου τῆς ἐν Κεγχρέαις.[15] In Jude, one of the group's most characteristic readings, v3u40-46.2, is shared by $\mathfrak{P}^{72}$, which may indicate ancient roots for the reading and possibly for the family. On further examination, however, this is the only significant group reading that $\mathfrak{P}^{72}$ supports apart from the much lower-weight reading in v4u20,[16] so the agreement is more likely coincidental.[17] In this case, the best explanation for the agreement is that the reading arose independently in $\mathfrak{P}^{72}$ and $\mathbf{f}^{618}$.

We will make some observations to conclude this section. While NMF in the uniform-weight setting succeeds at accounting for variance by giving more priority to common, widely divisive readings, NMF in the IDF setting does better at locating readings more exclusive to specific clusters. This, in turn, allows it to identify sparser reading bases and smaller MS groups.

For an illustration of this difference, please refer to Table 5, which gives a short list of widely-divided readings and their support among IDF-weighted NMF groups. Compare this to Table 3. Notice how splits among the groups formed tend to be more common for widely-divided readings in the IDF setting, while more distinctive group readings are identified in units with many readings, such as v5u12-20.

In general, the re-weighting of the observed data results in different clusters with sharper separation. While both weighting schemes isolate Alexandrian, $\mathbf{f}^{453}$, and lectionary clusters, IDF-weighted NMF compresses all previous subgroups of **K** into a single cluster, separates $\mathbf{f}^{1739}$ and $\mathbf{f}^{2138}$ appropriately, and identifies small subgroups of the Alexandrian and Byzantine texttypes with unique readings. For the purposes of identifying the most significant variation units and witnesses (e.g., to provide more refined and manageable inputs to CBGM or other more complex methods), this setting seems to be the most suitable.

---

[15]Specifically, the $\mathbf{f}^{618}$ MSS with the Romans subscription include **618**, **460**, **177**, **337** (which includes the subscription, but reorders many words and includes a reference to Tertius), **1738**, **607** (which omits πρὸς Ῥωμαίους), **1874** (which changes the begining to ἡ πρὸς Ῥωμαίους ἐπιστολή and omits τῆς ἐν Κεγχρέαις), and **404** (which appears to omit τῆς ἐν Κεγχρέαις).

[16]This unit concerns the inclusion or omission of the article τὸ, and all but a few witnesses (including most of $\mathbf{f}^{618}$) include the article, so this is probably an instance of independent errors producing agreement.

[17]Further evidence for coincidental agreement is that Wasserman lists $\mathfrak{P}^{72}$ as having a defective text reading here, with a corrector's hand supplying not the reading of $\mathbf{f}^{618}$, but the reading shared by the majority of MSS. Besides this, in the other high-ranking variation units of $\mathbf{f}^{618}$, $\mathfrak{P}^{72}$ either shares the majority reading or has a unique reading whose derivation from the $\mathbf{f}^{618}$ reading would be difficult to explain (e.g., in v5u12-20, v24u8-14, and v12u18). In other units, $\mathfrak{P}^{72}$ alternatively adds to (v10u24), omits from (v14u22-24), or transposes (v18u24-34) the $\mathbf{f}^{618}$ reading, leaving us no indication that the scribe of $\mathfrak{P}^{72}$ deviated from the $\mathbf{f}^{618}$ text in any consistent way.

Table 5. IDF-weight NMF cluster support for highly divided readings. Cluster-wide support is determined as in Table 3, and the readings are split as they are in Table 3.

|          | *Alex* | $\mathbf{f}^{453}$ | $\mathbf{f}^{1739}$ | $\mathbf{f}^{2138}$ | *Lect* | *Byz* | $\mathbf{f}^{915}$ | $\mathbf{f}^{618}$ |
|----------|--------|-------|--------|--------|--------|-------|-------|-------|
| v1u4-8   | I.X.   | I.X.  | I.X.   | I.X.   | X.I.   | Split | I.X.  | Split |
| v5u12-20 | Split  | ατκ.  | πΙα.   | πκα.   | Split  | Split | Split | υατκ. |
| v9u24-28 | Μωυσ.  | Μωυσ. | Μωσ.   | Μωσ.   | Split  | Split | Μωυσ. | Μωυσ. |
| v12u42-46| om.    | και   | om.    | om.    | και    | om.   | om.   | om.   |
| v15u14-18| om.    | om.   | om.    | om.    | om.    | αυτ.  | om.   | αυτ.  |
| v16u14-16| αυτ.   | Split | Split  | αυτ.   | Split  | Split | Split | Split |

## 6. Summary and Conclusions

In this paper, we have shown how non-negative matrix factorization, or NMF, can effectively classify MSS and readings on texttype-based principles. While, as a pre-genealogical method, it cannot make inferences regarding prior and posterior textual relationships, it can be used to facilitate more complex genealogical methods by providing better selections of readings and witnesses for input. We have demonstrated the suitability of NMF for these tasks on both theoretical and empirical grounds.

On the theoretical side, NMF is able to cluster both readings and MSS by finding the best approximate factorization of the collation matrix. By alternatively optimizing the basis and coefficient factor matrices, it keeps the problem of codependence between readings and MSS under control. Since, for certain NMF update rules, this process can be proven to stop only when the objective function of the factorization reaches a critical point, we have a theoretical guarantee that co-dependence won't cause an endless loop or result in arbitrarily bad clusters. Additionally, NMF provides an easy-to-interpret model that allows us to determine reading profiles for clusters and account for mixture between different clusters in MSS.

On the practical side, NMF provides fast, human-recognizable results for matrices in various weighted settings. NMF is able to factor a complete collation matrix of Jude for 518 MSS in less than 2 minutes in the uniform-weight setting, and in a matter of seconds in the IDF setting. As we have indicated, the uniform-weight setting provides good group classifications over widely-divided variation units, while the IDF-weight setting highlights the most distinctive readings of textual families.

Using NMF on Wasserman's collation of Jude, we were able to classify many previously-unclassified MSS and verify several existing group classifications. In the uniform-weight setting, we identified a distinct textual family for lectionaries in Jude; we found further empirical justification for von Soden's $\mathbf{K}^{r}$ and $\mathbf{K}^{c}$ groups, in addition to a subdivision of his $\mathbf{K}$ group; and we both verified the choices for the textual and marginal readings of Jude in [47] and proposed additional marginal readings based on the readings of the identified Byzantine subgroups. Meanwhile, in the IDF setting, we isolated characteristic MSS and readings for well-known groups, including the Alexandrian texttype, the Byzantine texttype, the lectionaries, $\mathbf{f}^{453}$, $\mathbf{f}^{1739}$, and $\mathbf{f}^{2138}$; we found two other groups, which we identified as $\mathbf{f}^{915}$ and $\mathbf{f}^{618}$; and we demonstrated that both $\mathbf{f}^{915}$ and $\mathbf{f}^{618}$ have family ties outside of the catholic epistles, suggesting that the classification made by NMF is legitimate.

We feel that NMF has tremendous potential as a tool for automatic, unsupervised, texttype-based textual criticism, and we wish to see it implemented in further studies. As we have attempted to show, its results can be fruitful in a multitude of applications, from organizing known collation data to performing exploratory analysis on what is yet unknown. We hope to find the new textual groupings and MS classifications done with NMF examined further and perhaps used as starting points for new research on the complex text of the epistle of Jude. It certainly deserves our greatest effort.

## Appendix A. Classification of Lacunose MSS

In Section 4.1, we explained that in the process of data selection, we regarded the texts of correctors and witnesses with fewer than 300 readings as non-continuous and therefore secondary. Table 6 lists the MSS from Wasserman's collation that were too lacunose to be included. Because of their age, most papyri and uncials are so lacunose that they must be excluded in this way. This leaves us with an unfortunate situation, in which we have nothing to say about the MSS in which we are most interested.

Fortunately, we are not without a remedy. Once NMF on the primary set of continuous-text witnesses has produced a basis matrix $\mathbf{W}$ for cluster readings, we can use this matrix to classify the secondary witnesses by whatever readings they do have. If we take $\vec{x}$ to be a vector representing the readings of a single secondary witness that we want to classify, then the solution $\vec{h}$ to the least-squares equation

$$(6) \qquad \arg\min_{\vec{h} \geq 0} ||\vec{x} - \mathbf{W}\vec{h}||_F^2$$

Table 6. MSS with fewer than 300 readings excluded from the primary NMF run.

| | | | | | | | |
|---|---|---|---|---|---|---|---|
| $\mathfrak{P}^{74}$ | $\mathfrak{P}^{78}$ | 025 | 0251 | 0316 | 8a | 69 | 172 |
| 602 | 610 | 613 | 614 | 712 | 720 | 832 | 913 |
| 1106 | 1360 | 1384 | 1724 | 1799 | 1831 | 1831S | 1852 |
| 1867 | 1895 | 1899 | 2138 | 2289 | 2356 | 2378 | 2511 |
| 2627 | 2653 | 2822 | ℓ6 | ℓ156 | ℓ427 | ℓ585 | ℓ617 |
| ℓ1281 | ℓ1818 | | | | | | |

will contain the mixture coefficients for the witness represented by $\vec{x}$. Because the entries of $\vec{h}$ are required to be non-negative, we can interepret $\vec{h}$ as we interpreted the mixture matrix $\mathbf{H}$ for the primary witnesses.

To solve equation 6, we used SciPy's `optimize.nnls` method for each MS in Table 6 individually, in both the uniform-weight case and the IDF-weight case. For the sake of space, we will not list the mixture coefficients of all MSS, but we will list the results for MS 2138 and the consistently-cited NA[28] witnesses $\mathfrak{P}^{74}$, $\mathfrak{P}^{78}$, 025, and 1852. These results can be found in Tables 7 and 8.

Table 7. Uniform-weight mixture coefficients for selected secondary MSS.

| | C1 | C2 | C3 | C4 | C5 | C6 | C7 | C8 |
|---|---|---|---|---|---|---|---|---|
| $\mathfrak{P}^{74}$ | 0.0470 | 0.0000 | 0.0000 | 0.0000 | 0.0000 | 0.0000 | 0.0000 | 0.0000 |
| $\mathfrak{P}^{78}$ | 0.0622 | 0.0379 | 0.0000 | 0.0000 | 0.0000 | 0.0000 | 0.0000 | 0.0000 |
| 025 | 0.4231 | 0.0000 | 0.2606 | 0.2068 | 0.0000 | 0.0365 | 0.0013 | 0.0383 |
| 1852 | 0.1967 | 1.1046 | 0.0000 | 0.0251 | 0.0000 | 0.0000 | 0.0000 | 0.0000 |
| 2138 | 0.2875 | 1.2018 | 0.0000 | 0.0000 | 0.0420 | 0.0000 | 0.0000 | 0.0000 |

Table 8. IDF-weight mixture coefficients for selected secondary MSS.

| | C1 | C2 | C3 | C4 | C5 | C6 | C7 | C8 |
|---|---|---|---|---|---|---|---|---|
| $\mathfrak{P}^{74}$ | 0.0000 | 0.0010 | 0.0000 | 0.0001 | 0.0001 | 0.0002 | 0.0000 | 0.0001 |
| $\mathfrak{P}^{78}$ | 0.1180 | 0.0026 | 0.0000 | 0.0000 | 0.0000 | 0.0171 | 0.0000 | 0.0000 |
| 025 | 0.0000 | 0.0815 | 0.0000 | 0.0000 | 0.0000 | 0.0189 | 0.0000 | 0.0006 |
| 1852 | 0.6950 | 0.0000 | 0.0537 | 0.2864 | 0.0000 | 0.0000 | 0.0561 | 0.0000 |
| 2138 | 1.6020 | 0.0000 | 0.0028 | 0.0000 | 0.0000 | 0.0084 | 0.0000 | 0.0339 |

As Table 7 shows, $\mathfrak{P}^{74}$ and $\mathfrak{P}^{78}$ are too lacunose to be classified with much confidence in the uniform-weight case. Uncial 025 (**P**) fares slightly better, exhibiting a moderate Alexandrian element and weaker lectionary and $\mathbf{K^r}$ elements. MSS 1852 and 2138, on the other hand, clearly belong to the $\mathbf{f}^{1739\,+\,2138}$ cluster.

The situation is not too different in the IDF-weighted setting, as Table 8 shows. As in the uniform-weight case, $\mathfrak{P}^{74}$'s extant readings do not commend it to any cluster. The situation is nearly the same for $\mathfrak{P}^{78}$; it has a slightly higher mixture coefficient for the $\mathbf{f}^{2138}$ cluster because it shares the group reading ἐπέχουσαι at v7u50, but it is otherwise too lacunose for us to determine any closer relationship. Uncial 025, like $\mathfrak{P}^{78}$, is inconclusive. MS 1852 shares characteristic readings of both $\mathbf{f}^{2138}$ and the Alexandrian cluster. Finally, as we would expect, MS 2138 clearly belongs to the family bearing its name.

## Appendix B. NMF Results

Table 9. Uniform-weight mixture coefficients for NA[28] consistently-cited MSS.

| | C1 | C2 | C3 | C4 | C5 | C6 | C7 | C8 |
|---|---|---|---|---|---|---|---|---|
| $\mathfrak{P}^{72}$ | 2.7600 | 0.0000 | 0.0000 | 0.0000 | 0.1153 | 0.0000 | 0.0000 | 0.0000 |
| 01 | 2.3008 | 0.4431 | 0.0000 | 0.0000 | 0.0000 | 0.0000 | 0.0971 | 0.1185 |
| 02 | 2.1142 | 0.5697 | 0.0000 | 0.0401 | 0.0000 | 0.0000 | 0.1965 | 0.0000 |
| 03 | 2.9769 | 0.0306 | 0.0000 | 0.0102 | 0.0000 | 0.0000 | 0.1604 | 0.0364 |
| 04 | 1.4799 | 0.9720 | 0.1648 | 0.0000 | 0.0000 | 0.0000 | 0.0000 | 0.0000 |
| 044 | 2.1660 | 0.3713 | 0.0000 | 0.0000 | 0.0000 | 0.0000 | 0.3413 | 0.0000 |
| 5 | 2.8403 | 0.1455 | 0.0102 | 0.0000 | 0.0000 | 0.0000 | 0.0000 | 0.0000 |
| 33 | 2.9074 | 0.2079 | 0.0000 | 0.0000 | 0.0000 | 0.1104 | 0.0000 | 0.0000 |
| 81 | 2.9994 | 0.1508 | 0.0000 | 0.0000 | 0.0000 | 0.1540 | 0.0204 | 0.0000 |
| 88 | 1.0954 | 0.7551 | 0.0000 | 0.0000 | 0.0000 | 0.2813 | 0.0982 | 0.2768 |
| 307 | 0.1903 | 0.0000 | 0.0000 | 0.0000 | 0.0000 | 2.0426 | 0.0000 | 0.0000 |
| 436 | 2.3373 | 0.0000 | 0.0000 | 0.2223 | 0.0000 | 0.3008 | 0.0000 | 0.0000 |
| 442 | 2.5224 | 0.4156 | 0.0000 | 0.0000 | 0.0000 | 0.2013 | 0.0000 | 0.0000 |
| 642 | 1.0912 | 0.0000 | 0.5090 | 0.1050 | 0.0000 | 0.1142 | 0.1693 | 0.3749 |
| 1175 | 0.0067 | 0.0121 | 0.0000 | 0.0336 | 1.1548 | 0.0000 | 0.0000 | 0.5077 |
| 1243 | 1.7896 | 0.9291 | 0.1265 | 0.0000 | 0.0000 | 0.0000 | 0.0000 | 0.0000 |
| 1448 | 0.3482 | 0.8453 | 0.0000 | 0.3396 | 0.4163 | 0.0000 | 0.0000 | 0.0000 |
| 1611 | 0.3658 | 2.0914 | 0.0000 | 0.0000 | 0.0000 | 0.0000 | 0.0000 | 0.0000 |
| 1735 | 2.3581 | 0.2282 | 0.0000 | 0.0000 | 0.3202 | 0.0000 | 0.0000 | 0.0000 |
| 1739 | 0.7424 | 1.7762 | 0.0013 | 0.0099 | 0.0000 | 0.0000 | 0.0000 | 0.0143 |
| 2344 | 2.3752 | 0.2548 | 0.0000 | 0.0000 | 0.0000 | 0.4282 | 0.0000 | 0.0000 |
| 2492 | 0.7405 | 0.3525 | 0.2230 | 0.3013 | 0.0906 | 0.0333 | 0.0202 | 0.2314 |

Table 10. Uniform-weight cluster 1
MSS, sorted by $\mathbf{H}_{1,j}$.

| ms | $\mathbf{H}_{1,j}$ |
|---:|---:|
| 81 | 2.9994 |
| 623 | 2.9977 |
| 03 | 2.9769 |
| 326 | 2.9536 |
| 33 | 2.9074 |
| 5 | 2.8403 |
| 1837 | 2.7880 |
| $\mathfrak{P}^{72}$ | 2.7600 |
| $\ell$596 | 2.6330 |
| 93 | 2.5811 |
| 442 | 2.5224 |
| 1845 | 2.5080 |
| 2805 | 2.4686 |
| 1409 | 2.3977 |
| 665 | 2.3766 |

Table 11. Uniform-weight cluster 2
MSS, sorted by $\mathbf{H}_{2,j}$.

| ms | $\mathbf{H}_{2,j}$ |
|---:|---:|
| 1505 | 2.1731 |
| 2495 | 2.1515 |
| 1292 | 2.1169 |
| 1611 | 2.0914 |
| 630 | 1.9175 |
| 322 | 1.8994 |
| 1241 | 1.8967 |
| 2298 | 1.8948 |
| 323 | 1.8877 |
| 1739 | 1.7762 |
| 2200 | 1.7431 |
| 1881 | 1.6618 |
| 1765 | 1.2892 |
| 1832 | 1.2892 |
| 2494 | 1.2726 |

Table 12. Uniform-weight cluster 3
MSS, sorted by $\mathbf{H}_{3,j}$.

| ms | $\mathbf{H}_{3,j}$ |
|---:|---:|
| $\ell$606 | 1.8747 |
| $\ell$938 | 1.8747 |
| $\ell$840 | 1.8572 |
| $\ell$740 | 1.8396 |
| $\ell$145 | 1.8388 |
| $\ell$2106 | 1.8300 |
| $\ell$809 | 1.8280 |
| $\ell$2394 | 1.8264 |
| $\ell$62 | 1.7966 |
| $\ell$604 | 1.7849 |
| $\ell$1279 | 1.7678 |
| $\ell$623 | 1.7337 |
| $\ell$1141 | 1.6346 |
| $\ell$921 | 1.6124 |
| $\ell$162 | 1.6005 |

Table 13. Uniform-weight cluster 4
MSS, sorted by $\mathbf{H}_{4,j}$.

| ms | $\mathbf{H}_{4,j}$ |
|---:|---:|
| 141 | 1.1874 |
| 204 | 1.1874 |
| 394 | 1.1874 |
| 444 | 1.1874 |
| 1101 | 1.1874 |
| 1723 | 1.1874 |
| 1737 | 1.1874 |
| 1752 | 1.1874 |
| 1865 | 1.1874 |
| 2221 | 1.1874 |
| 2431 | 1.1874 |
| 2554 | 1.1874 |
| 2723 | 1.1874 |
| 2255 | 1.1865 |
| 1748 | 1.1819 |

TABLE 14. Uniform-weight cluster 5
MSS, sorted by $\mathbf{H}_{5,j}$.

| ms | $\mathbf{H}_{5,j}$ |
|---:|---|
| 1769 | 1.4799 |
| 1780 | 1.4703 |
| 2705 | 1.4236 |
| 451 | 1.3881 |
| 2516 | 1.3875 |
| 1424 | 1.3575 |
| 330 | 1.3500 |
| 1734 | 1.3397 |
| 619 | 1.2821 |
| 302 | 1.2384 |
| 637 | 1.2238 |
| 325 | 1.2176 |
| 049 | 1.1871 |
| 627 | 1.1752 |
| 601 | 1.1586 |

TABLE 15. Uniform-weight cluster 6
MSS, sorted by $\mathbf{H}_{6,j}$.

| ms | $\mathbf{H}_{6,j}$ |
|---:|---|
| 321 | 2.1357 |
| 918 | 2.0686 |
| 2197 | 2.0535 |
| 307 | 2.0426 |
| 453 | 2.0384 |
| 2818 | 1.9292 |
| 1678 | 1.7791 |
| 94 | 1.6145 |
| 2186 | 1.4535 |
| 1840 | 1.3808 |
| 378 | 1.3300 |
| 2652 | 1.3234 |
| 2147 | 1.3211 |
| 1838 | 1.1785 |
| 629 | 1.1530 |

TABLE 16. Uniform-weight cluster 7
MSS, sorted by $\mathbf{H}_{7,j}$.

| ms | $\mathbf{H}_{7,j}$ |
|---:|---|
| 056 | 1.5676 |
| 639 | 1.5446 |
| 1066 | 1.5190 |
| 0142 | 1.5151 |
| 641 | 1.3858 |
| 327 | 1.3669 |
| 606 | 1.3596 |
| 1352 | 1.3558 |
| 103 | 1.3384 |
| 1853 | 1.2370 |
| 312 | 1.2364 |
| 218 | 1.1932 |
| 1103 | 1.1730 |
| 454 | 1.1665 |
| 452 | 1.1517 |

TABLE 17. Uniform-weight cluster 8
MSS, sorted by $\mathbf{H}_{8,j}$.

| ms | $\mathbf{H}_{8,j}$ |
|---:|---|
| 912 | 1.9451 |
| 390 | 1.9369 |
| 1863 | 1.9369 |
| 1861 | 1.9295 |
| 2085 | 1.9256 |
| 234 | 1.8782 |
| 1753 | 1.8289 |
| 2279 | 1.8195 |
| 42 | 1.7922 |
| 1003 | 1.7062 |
| 996 | 1.6769 |
| 1594 | 1.6531 |
| 1727 | 1.6099 |
| 1661 | 1.5985 |
| 1405 | 1.5814 |

TABLE 18. Uniform-weight cluster 1 readings, sorted by $\mathbf{W}_{i,1}$.

| reading | $\mathbf{W}_{i,1}$ |
|---|---|
| v9u24-28.1 | 0.3925 |
| v25u32-38.1 | 0.3638 |
| v17u12-16.1 | 0.3381 |
| v13u30-34.1 | 0.3240 |
| v12u42-46.1 | 0.3175 |
| v25u10-20.1 | 0.3172 |
| v9u36.1 | 0.3164 |
| v1u16-22.1 | 0.3090 |
| v18u6.1 | 0.3079 |
| v4u32.1 | 0.3064 |
| v2u6.1 | 0.3048 |
| v8u4.1 | 0.3034 |
| v24u2-6.1 | 0.3019 |
| v3u26-34.1 | 0.2999 |
| v10u20.1 | 0.2998 |

TABLE 19. Uniform-weight cluster 2 readings, sorted by $\mathbf{W}_{i,2}$.

| reading | $\mathbf{W}_{i,2}$ |
|---|---|
| v15u45.2 | 0.5917 |
| v17u12-16.8 | 0.5821 |
| v19u8.3 | 0.5122 |
| v15u12.1 | 0.4992 |
| v9u24-28.2 | 0.4797 |
| v25u10-20.1 | 0.4652 |
| v12u18.1 | 0.4608 |
| v15u32-34.1 | 0.4596 |
| v15u10.1 | 0.4576 |
| v14u22-24.1 | 0.4571 |
| v2u6.1 | 0.4564 |
| v11u4.1 | 0.4559 |
| v15u10-20.1 | 0.4545 |
| v15u10-16.1 | 0.4545 |
| v18u6.1 | 0.4526 |

TABLE 20. Uniform-weight cluster 3 readings, sorted by $\mathbf{W}_{i,3}$.

| reading | $\mathbf{W}_{i,3}$ |
|---|---|
| v19u8.3 | 0.6279 |
| v12u42-46.7 | 0.5950 |
| v1u4-8.2 | 0.5833 |
| v2u6.1 | 0.5817 |
| v9u24-28.2 | 0.5777 |
| v18u6.1 | 0.5572 |
| v20u26.1 | 0.5571 |
| v2u8-12.1 | 0.5564 |
| v15u10.1 | 0.5550 |
| v13u30-34.3 | 0.5526 |
| v15u10-20.1 | 0.5513 |
| v15u10-16.1 | 0.5513 |
| v15u12.1 | 0.5508 |
| v12u30-32.1 | 0.5495 |
| v6u4.1 | 0.5479 |

TABLE 21. Uniform-weight cluster 4 readings, sorted by $\mathbf{W}_{i,4}$.

| reading | $\mathbf{W}_{i,4}$ |
|---|---|
| v15u14-18.2 | 0.9035 |
| v23u2-22.15 | 0.9016 |
| v21u2-10.1 | 0.9001 |
| v16u14-16.1 | 0.8925 |
| v4u48-58.5 | 0.8863 |
| v20u26.1 | 0.8862 |
| v12u19.1 | 0.8738 |
| v18u6.1 | 0.8696 |
| v1u24.2 | 0.8666 |
| v12u6.2 | 0.8651 |
| v19u8.1 | 0.8638 |
| v3u18-22.7 | 0.8623 |
| v15u20-30.1 | 0.8609 |
| v25u3.2 | 0.8604 |
| v6u20.1 | 0.8597 |

TABLE 22. Uniform-weight cluster 5 readings, sorted by $\mathbf{W}_{i,5}$.

| reading | $\mathbf{W}_{i,5}$ |
|---|---|
| v1u26-34.3 | 0.8233 |
| v19u8.1 | 0.7554 |
| v5u12-20.17 | 0.7411 |
| v23u2-22.15 | 0.7061 |
| v4u48-58.5 | 0.7029 |
| v15u14-18.2 | 0.6880 |
| v18u24-34.1 | 0.6734 |
| v24u8-14.6 | 0.6673 |
| v15u6-8.1 | 0.6599 |
| v1u4-8.2 | 0.6552 |
| v20u8-18.2 | 0.6487 |
| v15u45.1 | 0.6476 |
| v15u1.1 | 0.6429 |
| v12u19.1 | 0.6419 |
| v9u44-46.1 | 0.6419 |

TABLE 23. Uniform-weight cluster 6 readings, sorted by $\mathbf{W}_{i,6}$.

| reading | $\mathbf{W}_{i,6}$ |
|---|---|
| v12u42-46.7 | 0.5715 |
| v13u8.2 | 0.5495 |
| v25u24-30.3 | 0.5445 |
| v1u24.2 | 0.5334 |
| v25u3.2 | 0.5236 |
| v25u10-20.1 | 0.5127 |
| v20u26.1 | 0.5117 |
| v9u44-46.1 | 0.5101 |
| v16u14-16.3 | 0.5080 |
| v24u18.1 | 0.5015 |
| v2u6.1 | 0.4959 |
| v15u1.1 | 0.4952 |
| v7u24-28.2 | 0.4946 |
| v11u4.1 | 0.4912 |
| v18u6.1 | 0.4889 |

TABLE 24. Uniform-weight cluster 7 readings, sorted by $\mathbf{W}_{i,7}$.

| reading | $\mathbf{W}_{i,7}$ |
|---|---|
| v16u14-16.3 | 0.8088 |
| v13u30-34.3 | 0.7319 |
| v12u19.1 | 0.7083 |
| v12u6.2 | 0.6222 |
| v25u4-8.1 | 0.6160 |
| v9u48.1 | 0.6069 |
| v25u3.2 | 0.5936 |
| v3u18-22.7 | 0.5887 |
| v6u20.1 | 0.5785 |
| v24u18.1 | 0.5783 |
| v22u2-10.3 | 0.5771 |
| v11u1.1 | 0.5730 |
| v25u10-20.6 | 0.5699 |
| v11u4.1 | 0.5684 |
| v9u44-46.1 | 0.5681 |

TABLE 25. Uniform-weight cluster 8 readings, sorted by $\mathbf{W}_{i,8}$.

| reading | $\mathbf{W}_{i,8}$ |
|---|---|
| v9u24-28.1 | 0.7138 |
| v1u4-8.2 | 0.5856 |
| v20u26.1 | 0.5769 |
| v15u14-18.2 | 0.5759 |
| v2u6.1 | 0.5724 |
| v21u2-10.1 | 0.5661 |
| v3u18-22.7 | 0.5593 |
| v1u26-34.1 | 0.5577 |
| v8u24-28.1 | 0.5535 |
| v9u36.1 | 0.5510 |
| v16u14-16.3 | 0.5503 |
| v13u30-34.1 | 0.5482 |
| v18u10-18.17 | 0.5466 |
| v15u20-30.1 | 0.5415 |
| v25u10-20.6 | 0.5392 |

TABLE 26. IDF-weight mixture coefficients for NA$^{28}$ consistently-cited MSS.

| | C1 | C2 | C3 | C4 | C5 | C6 | C7 | C8 |
|---|---|---|---|---|---|---|---|---|
| $\mathfrak{P}^{72}$ | 0.0000 | 0.0000 | 0.0000 | 0.9034 | 0.0000 | 0.0000 | 0.0078 | 0.2398 |
| 01 | 0.1628 | 0.0000 | 0.1149 | 0.6247 | 0.0000 | 0.0000 | 0.2579 | 0.0559 |
| 02 | 0.0787 | 0.0000 | 0.0243 | 0.9383 | 0.0000 | 0.0000 | 0.0142 | 0.0499 |
| 03 | 0.0842 | 0.0000 | 0.0000 | 0.7854 | 0.0000 | 0.0000 | 0.1755 | 0.0000 |
| 04 | 0.4510 | 0.0000 | 0.2242 | 0.3155 | 0.0000 | 0.0486 | 0.2068 | 0.0448 |
| 044 | 0.0000 | 0.0674 | 0.0944 | 0.7535 | 0.0000 | 0.0000 | 0.1762 | 0.0000 |
| 5 | 0.0420 | 0.0030 | 0.0000 | 0.8648 | 0.0193 | 0.0266 | 0.0017 | 0.0000 |
| 33 | 0.0000 | 0.0000 | 0.0517 | 0.9173 | 0.0000 | 0.0000 | 0.0000 | 0.0614 |
| 81 | 0.0000 | 0.0000 | 0.1123 | 0.8761 | 0.1115 | 0.0000 | 0.0000 | 0.0000 |
| 88 | 0.0000 | 0.0174 | 0.0251 | 0.0000 | 0.0000 | 0.0000 | 1.6187 | 0.0000 |
| 307 | 0.0000 | 0.0000 | 0.0000 | 0.0000 | 1.1342 | 0.0000 | 0.0000 | 0.0000 |
| 436 | 0.0000 | 0.0114 | 0.0038 | 0.6146 | 0.1816 | 0.0481 | 0.0386 | 0.0673 |
| 442 | 0.1218 | 0.0000 | 0.0000 | 0.4513 | 0.1147 | 0.0000 | 0.9392 | 0.0000 |
| 642 | 0.0000 | 0.0697 | 0.0000 | 0.2661 | 0.0763 | 0.1628 | 0.0000 | 0.0000 |
| 1175 | 0.0000 | 0.3115 | 0.0000 | 0.0092 | 0.0000 | 0.0000 | 0.0000 | 0.1045 |
| 1243 | 0.6215 | 0.0000 | 0.1516 | 0.3224 | 0.0000 | 0.1095 | 0.6035 | 0.0065 |
| 1448 | 0.5217 | 0.1217 | 0.0646 | 0.1592 | 0.0000 | 0.0000 | 0.0561 | 0.0000 |
| 1611 | 1.9718 | 0.0000 | 0.0090 | 0.1460 | 0.0000 | 0.0000 | 0.0000 | 0.0066 |
| 1735 | 0.2203 | 0.1801 | 0.0000 | 0.7345 | 0.0000 | 0.0000 | 0.0000 | 0.0000 |
| 1739 | 0.0015 | 0.0107 | 1.4197 | 0.0710 | 0.0000 | 0.0000 | 0.0100 | 0.0000 |
| 2344 | 0.0378 | 0.0000 | 0.0638 | 0.8365 | 0.1816 | 0.0000 | 0.0000 | 0.0000 |
| 2492 | 0.0941 | 0.0382 | 0.1596 | 0.0075 | 0.0000 | 0.0917 | 0.2643 | 0.0804 |

Table 27. IDF-weight cluster 1 MSS, sorted by $\mathbf{H}_{1,j}$.

| ms | $\mathbf{H}_{1,j}$ |
|---:|---|
| 1505 | 2.3097 |
| 2495 | 2.1089 |
| 1611 | 1.9718 |
| 1292 | 1.8218 |
| 630 | 1.7394 |
| 2200 | 1.6920 |
| 1765 | 1.3502 |
| 1832 | 1.3502 |
| 2494 | 1.3313 |
| 876 | 1.2328 |
| 2412 | 0.9021 |
| 2147 | 0.8925 |
| 2652 | 0.8744 |
| 2243 | 0.8615 |
| 378 | 0.7244 |

Table 28. IDF-weight cluster 2 MSS, sorted by $\mathbf{H}_{2,j}$.

| ms | $\mathbf{H}_{2,j}$ |
|---:|---|
| 454 | 0.4207 |
| 641 | 0.4132 |
| 2125 | 0.4103 |
| 606 | 0.4072 |
| 221 | 0.4006 |
| 625 | 0.3858 |
| 250 | 0.3789 |
| 1888 | 0.3742 |
| 103 | 0.3685 |
| 018 | 0.3675 |
| 393 | 0.3566 |
| 314 | 0.3529 |
| 309 | 0.3453 |
| 616 | 0.3451 |
| 1862 | 0.3445 |

Table 29. IDF-weight cluster 3 MSS, sorted by $\mathbf{H}_{3,j}$.

| ms | $\mathbf{H}_{3,j}$ |
|---:|---|
| 323 | 1.4883 |
| 1241 | 1.4483 |
| 1739 | 1.4197 |
| 322 | 1.4163 |
| 1881 | 1.3174 |
| 2298 | 1.2088 |
| 6 | 0.8429 |
| 1501 | 0.3657 |
| 2805 | 0.2927 |
| 2374 | 0.2308 |
| 93 | 0.2258 |
| 04 | 0.2242 |
| 665 | 0.2177 |
| 2492 | 0.1596 |
| 1243 | 0.1516 |

Table 30. IDF-weight cluster 4 MSS, sorted by $\mathbf{H}_{4,j}$.

| ms | $\mathbf{H}_{4,j}$ |
|---:|---|
| 326 | 0.9914 |
| 623 | 0.9607 |
| 93 | 0.9568 |
| 1837 | 0.9551 |
| 02 | 0.9383 |
| 61 | 0.9372 |
| 665 | 0.9225 |
| 33 | 0.9173 |
| 𝔓⁷² | 0.9034 |
| 81 | 0.8761 |
| 5 | 0.8648 |
| 2344 | 0.8365 |
| 03 | 0.7854 |
| 044 | 0.7535 |
| 2805 | 0.7425 |

Table 31. IDF-weight cluster 5 MSS, sorted by $\mathbf{H}_{5,j}$.

| ms | $\mathbf{H}_{5,j}$ |
|---:|---:|
| 918 | 1.2273 |
| 321 | 1.1608 |
| 453 | 1.1342 |
| 307 | 1.1342 |
| 2197 | 1.0957 |
| 2818 | 1.0141 |
| 94 | 0.9844 |
| 378 | 0.9693 |
| 1678 | 0.9176 |
| 2147 | 0.8489 |
| 2652 | 0.7936 |
| 2412 | 0.5837 |
| 2186 | 0.5343 |
| 1840 | 0.5193 |
| 629 | 0.4642 |

Table 32. IDF-weight cluster 6 MSS, sorted by $\mathbf{H}_{6,j}$.

| ms | $\mathbf{H}_{6,j}$ |
|---:|---:|
| ℓ145 | 0.7782 |
| ℓ1279 | 0.7631 |
| ℓ604 | 0.7601 |
| ℓ2394 | 0.7354 |
| ℓ62 | 0.7190 |
| ℓ606 | 0.7042 |
| ℓ938 | 0.7042 |
| ℓ921 | 0.7007 |
| ℓ740 | 0.6648 |
| ℓ840 | 0.6409 |
| ℓ1505 | 0.6287 |
| ℓ809 | 0.6270 |
| ℓ2106 | 0.6270 |
| ℓ1141 | 0.6253 |
| ℓ623 | 0.6077 |

Table 33. IDF-weight cluster 7 MSS, sorted by $\mathbf{H}_{7,j}$.

| ms | $\mathbf{H}_{7,j}$ |
|---:|---:|
| 915 | 1.7575 |
| 88 | 1.6187 |
| 1846 | 1.4810 |
| 459 | 1.2416 |
| 1845 | 1.1819 |
| 104 | 1.0390 |
| 1838 | 1.0124 |
| 1842 | 0.9582 |
| 442 | 0.9392 |
| ℓ596 | 0.8555 |
| 621 | 0.8114 |
| 1243 | 0.6035 |
| 181 | 0.3188 |
| 1836 | 0.2770 |
| 2492 | 0.2643 |

Table 34. IDF-weight cluster 8 MSS, sorted by $\mathbf{H}_{8,j}$.

| ms | $\mathbf{H}_{8,j}$ |
|---:|---:|
| 618 | 1.4430 |
| 460 | 1.3495 |
| 177 | 1.3084 |
| 337 | 1.2419 |
| 1738 | 1.2167 |
| 180 | 0.3425 |
| 1875 | 0.3401 |
| 607 | 0.3385 |
| 390 | 0.2978 |
| 1863 | 0.2978 |
| 1753 | 0.2832 |
| 234 | 0.2711 |
| 912 | 0.2694 |
| 2675 | 0.2646 |
| 2279 | 0.2588 |

TABLE 35. IDF-weight cluster 1 readings, sorted by $\mathbf{W}_{i,1}$.

| reading | $\mathbf{W}_{i,1}$ |
|---|---|
| v2u8-12.5 | 0.8484 |
| v15u20-30.9 | 0.8306 |
| v1u26-34.13 | 0.7902 |
| v5u12-20.6 | 0.7463 |
| v7u50.3 | 0.7249 |
| v17u12-16.8 | 0.7218 |
| v1u16-22.8 | 0.6602 |
| v24u18.2 | 0.6266 |
| v18u10-18.1 | 0.5988 |
| v14u26-32.2 | 0.5956 |
| v9u48.2 | 0.5600 |
| v23u2-22.17 | 0.5473 |
| v25u24-30.1 | 0.5401 |
| v18u24-34.8 | 0.5318 |
| v1u24.1 | 0.5081 |

TABLE 36. IDF-weight cluster 2 readings, sorted by $\mathbf{W}_{i,2}$.

| reading | $\mathbf{W}_{i,2}$ |
|---|---|
| v21u2-10.16 | 0.9665 |
| v20u26.5 | 0.9311 |
| v1u26-34.3 | 0.9240 |
| inscriptio.7 | 0.8848 |
| v13u30-34.3 | 0.8579 |
| v23u2-22.15 | 0.7981 |
| subscriptio.1 | 0.7458 |
| v1u4-8.2 | 0.7452 |
| v16u14-16.1 | 0.7205 |
| v24u8-14.6 | 0.7124 |
| inscriptio.8 | 0.6934 |
| v9u24-28.2 | 0.6720 |
| v15u12.2 | 0.6591 |
| v9u36.2 | 0.6467 |
| v25u40-52.14 | 0.6298 |

TABLE 37. IDF-weight cluster 3 readings, sorted by $\mathbf{W}_{i,3}$.

| reading | $\mathbf{W}_{i,3}$ |
|---|---|
| v15u20-30.15 | 1.3171 |
| v15u14-18.5 | 1.2861 |
| v24u8-14.20 | 1.2779 |
| v9u48.3 | 1.2409 |
| v17u8.3 | 1.2354 |
| v5u12-20.8 | 1.2032 |
| v5u4.2 | 1.1579 |
| v14u4-8.3 | 1.1385 |
| v15u1.2 | 1.1079 |
| v9u44-46.5 | 1.0496 |
| v18u10-18.9 | 0.9143 |
| v17u12-16.8 | 0.8983 |
| v14u26-32.2 | 0.8918 |
| v22u2-10.4 | 0.8826 |
| v23u2-22.1 | 0.8479 |

TABLE 38. IDF-weight cluster 4 readings, sorted by $\mathbf{W}_{i,4}$.

| reading | $\mathbf{W}_{i,4}$ |
|---|---|
| v25u32-38.1 | 1.0976 |
| v7u24-28.1 | 1.0775 |
| v20u8-18.1 | 1.0383 |
| v22u2-10.4 | 0.9979 |
| v13u8.1 | 0.9335 |
| v25u24-30.1 | 0.8967 |
| v23u2-22.1 | 0.8250 |
| v25u3.1 | 0.8116 |
| v3u18-22.1 | 0.7924 |
| v24u8-14.1 | 0.7753 |
| v25u10-20.1 | 0.7200 |
| v15u14-18.1 | 0.7050 |
| v12u6.1 | 0.6736 |
| v14u26-32.8 | 0.6101 |
| v18u10-18.5 | 0.6076 |

TABLE 39. IDF-weight cluster 5 readings, sorted by $\mathbf{W}_{i,5}$.

| reading | $\mathbf{W}_{i,5}$ |
|---|---|
| v24u8-14.19 | 1.2565 |
| v23u2-22.8 | 1.1999 |
| v25u40-52.18 | 1.1713 |
| v25u32-38.2 | 1.0491 |
| v8u4.2 | 1.0478 |
| v5u12-20.26 | 0.9678 |
| v15u20-30.12 | 0.9245 |
| v20u8-18.10 | 0.8915 |
| v18u10-18.9 | 0.8658 |
| v15u45.2 | 0.7992 |
| v3u18-22.1 | 0.7974 |
| v12u6.1 | 0.7629 |
| v15u14-18.1 | 0.7266 |
| v25u10-20.1 | 0.7018 |
| v4u48-58.1 | 0.6862 |

TABLE 40. IDF-weight cluster 6 readings, sorted by $\mathbf{W}_{i,6}$.

| reading | $\mathbf{W}_{i,6}$ |
|---|---|
| v23u2-22.23 | 1.4580 |
| v11u4.4 | 1.3425 |
| v11u1.4 | 1.2776 |
| v12u19.2 | 1.2760 |
| v12u42-46.7 | 1.0131 |
| v19u8.3 | 1.0050 |
| v15u14-18.1 | 0.8002 |
| v18u10-18.30 | 0.6982 |
| v4u48-58.1 | 0.6840 |
| v1u4-8.2 | 0.6205 |
| v13u30-34.3 | 0.5402 |
| v4u48-58.9 | 0.4637 |
| v9u44-46.3 | 0.4491 |
| v25u4-8.12 | 0.3935 |
| v15u6-8.4 | 0.3857 |

TABLE 41. IDF-weight cluster 7 readings, sorted by $\mathbf{W}_{i,7}$.

| reading | $\mathbf{W}_{i,7}$ |
|---|---|
| v14u26-32.9 | 1.1798 |
| v22u2-10.1 | 0.9192 |
| v12u12.2 | 0.8706 |
| v6u20.2 | 0.8104 |
| v4u48-58.12 | 0.7723 |
| v5u12-20.11 | 0.6454 |
| v18u2-4.3 | 0.6310 |
| v9u49.2 | 0.6258 |
| v15u45.2 | 0.6081 |
| subscriptio.18 | 0.5921 |
| v25u40-52.5 | 0.5870 |
| v7u24-28.12 | 0.5654 |
| v12u6.1 | 0.5503 |
| v20u8-18.1 | 0.5447 |
| v23u2-22.10 | 0.5127 |

TABLE 42. IDF-weight cluster 8 readings, sorted by $\mathbf{W}_{i,8}$.

| reading | $\mathbf{W}_{i,8}$ |
|---|---|
| v18u24-34.2 | 1.2782 |
| v3u40-46.2 | 1.2004 |
| v10u24.2 | 1.1830 |
| v20u8-18.12 | 1.1667 |
| v5u12-20.15 | 1.1442 |
| v14u22-24.3 | 1.0127 |
| v24u8-14.7 | 0.9850 |
| v12u18.2 | 0.9393 |
| inscriptio.3 | 0.9303 |
| v12u12.2 | 0.8887 |
| subscriptio.13 | 0.7942 |
| v6u36-38.5 | 0.6020 |
| v4u48-58.9 | 0.4657 |
| v25u4-8.12 | 0.4266 |
| v9u24-28.1 | 0.3208 |